\documentclass{article}

\usepackage{hyperref}       
\usepackage{url}            
\usepackage{booktabs}       
\usepackage{amsfonts}       
\usepackage{nicefrac}       
\usepackage{graphicx}
\usepackage{natbib}
\usepackage{doi}
\usepackage{amsmath}

\usepackage{cancel}
\usepackage{bm}

\newcommand{\lp}[0]{\left(}
\newcommand{\rp}[0]{\right)}
\newcommand{\Q}[0]{\mathbf{Q}}
\newcommand{\z}[0]{\mathbf{z}}
\newcommand{\A}[0]{\mathbf{A}}
\newcommand{\bk}[0]{\mathbf{b}}

\title{A Factor Graph-based approach to vehicle sideslip angle estimation}

\author{Antonio Leanza $^{1}$, Giulio Reina $^{1}$ and Jos\'e-Luis Blanco-Claraco ${}^{2}$\\
~\\
$^{1}$ \quad Department of Mechanics, Mathematics, and Management,\\
Polytechnic of Bari, via Orabona 4, 70126 Bari, Italy\\
$^{2}$ \quad Department of Engineering, University of Almer\'ia, \\
Campus de Excelencia Internacional Agroalimentario, ceiA3,\\
04120 Almer\'ia, Spain
}

\begin{document}
\maketitle

\begin{abstract}
Sideslip angle is an important variable for understanding and monitoring vehicle dynamics
but it lacks an inexpensive method for direct measurement. 
Therefore, it is typically estimated from inertial and other proprioceptive sensors onboard 
using filtering methods from the family of the Kalman Filter.
As a novel alternative, this work proposes modelling the problem directly as a graphical model (factor graph),
which can then be optimized using a variety of methods, such as whole dataset batch optimization 
for offline processing or fixed-lag smoother for on-line operation.
Experimental results on real vehicle datasets validate the proposal with a good agreement between estimated and actual sideslip angle, showing similar performance than  the state-of-the-art with a great potential for future extensions due to the flexible mathematical framework.
\end{abstract}


%

\section{Introduction}

In the last 30 years a large amount of research and papers have been devoted to side-slip estimation, because it represents a fundamental feature for vehicle dynamics \cite{LenzoReview}. Despite its central role, a direct measurement of the side-slip angle value over time during vehicle motion is often impractical, due to technical and economic reasons. For instance, a cheap but few practical solution is to use measures coming from Global Position System (GPS) that can provide the position of the receiver without any numerical integration and then derive the velocity of the vehicle using Doppler measurements \cite{LenzoReview68}. Nevertheless, GPS receivers have intrinsic issues such as temporary signal unavailability due to surroundings such as trees, tall buildings, mountains and tunnels, as well as different working frequencies with respect to other sensors involved in vehicle dynamics control, such as accelerometers, gyroscopes and so on. On the other hand, a direct measurement of the vehicle side slip angle can be achieved by high precision optical sensors, but they are too much sophisticated, still in Research and Development (R\&D) stage and too much expensive to be suitable for production vehicles \cite{OpticalSensor}.

For these reasons, the side slip angle estimation continues to play a significant role in vehicle dynamics, attracting noticeable interest in the academic and industrial worlds \cite{LenzoReview2,Mastinu,LenzoReview4}. Several methods have been developed and described throughout the scientific literature, which make use of different models and estimators. The most common way relies on model-based observers \cite{LenzoReview13,LenzoReview15,LenzoReview36,LenzoReview49,LenzoReview30}, which make use of a vehicle reference model for state and parameters estimators. Different levels of sophistication can be obtained in order to achieve a phenomenon description as accurate as possible. One of the most common combinations that can be found in the literature is the bicycle (single-track) model as an observer for a Kalman Filter (KF). This arrangement allows the estimation of states and parameters in the same time. For instance in \cite{ReinaPaianoBlanco,ReinaMessina} a linear bicycle model is used in combination with an Extended Kalman Filter (EKF) for the estimation of the side slip angle and vehicle parameters in the same time. The former, estimates lateral velocity and vehicle mass, by correcting also the bias of the gyroscope, instead the latter estimates the front and rear cornering stiffness and the side slip angle directly as a dynamic state. \cite{CKFlinear} exploits the same linear model, but use a Cubature Kalman Filter (CKF) as estimator for obtaining the side slip angle and other states from common measures (i.e. accelerations, vehicle longitudinal velocity, etc.). In \cite{LenzoReview16} a combining approach between kinematic and dynamic model is carried out, since the kinematic model performs well during transient maneuvers but fails in the steady state conditions. Therefore, the information provided by the kinematic formulation are exploit to update the single-track model parameters (i.e. the tire cornering stiffness), while the dynamic state observer is used in nearly quasi-state condition. The steady state or transient conditions are discriminate via fuzzy-logic procedure. Furthermore, the bicycle model can be also coupled with nonlinear tire models, as in \cite{LeanzaVSD6}, where the authors aim to estimate the side slip angle of a heavy-duty vehicle by using a Rational tire model and an EKF as estimator. This study showed good estimation performance, but at the cost of an overparametrised and complex model. On the other hand, the four-contact models provide a better description of vehicle dynamics, but obviously at the cost of a greater number of parameters and an increase in the complexity of the systems. As instance, \cite{LenzoReview18} uses an EKF applied to a four-contact vehicle model with a Dugoff tire model in order to estimate the side slip angle and the the tire/road forces. \cite{LenzoReview21} uses again a four-contact model, but with a semiphysical non-linear tire model ``Unitire'' and applies a redeuced-order Sliding Mode Observer (SMO), evaluating the performance of the proposed method by means of simulation and experiments. In \cite{LenzoReview55} the Magic Formula (or Pacejka tire model \cite{pacejka2002tyre}) is exploited, where a preliminary filtering on vertical forces is performed via linear KF in order to estimate the roll angle and then an EKF is applied to the four-contact vehicle model to achieve the side slip angle. Again, Magic Formula with a four-contact vehicle model is used in \cite{LeanzaVSD8}, where a dual estimation scheme was adopted: the side slip angle and tire/road forces by means of a Dual Unscented Kalman Filter (UKF) algorithm and the Pacejka tire parameters by solving a nonlinear Least Squares (LS) problem. However, the sophistication of these models might be unpractical in real applications. To overcome this issue, some authors, as \cite{LeanzaReinaMantriota} rely on direct causality equations without the need of any explicit tire model. In that research different estimators are developed, as the standard EKF, the CKF and Particle filtering (PF) and the results are evaluated by using a vehicle model with 14 Degrees Of Freedom (DOFs) which performs standard maneuvers. 
The work \cite{LenzoReview58} performs vehicle sideslip angle and road bank angle estimation via simple algebraic relationship in real time, based on two online parameters identification techniques, combining single-track model with roll and tire slip models and force due to bank angle. Moreover, with the use of a lateral G sensor signal and by including road bank angle effect, the front and rear cornering stiffness and vehicle side slip angle are identified, in the absence of any a priori knowledge on the road bank angle. In order to completely overcome the need for a vehicle model of any kind and its related complex set of parameters, different approaches based on Artificial Neural Networks (ANN) \cite{LenzoReview101,LenzoReview106,LenzoReview96,LenzoReview102} are widely regarded nowadays, since they are suitable to model complex systems just using their ability to identify relationships from input–output data.

In this paper a novel estimation technique grounded in the Factor Graphs (FGs) theory is presented. To the best of authors' knowledge, it is the first time that a FG-based observer is proposed in the automotive field. A FG is a type of probabilistic graphical model that can be used to describe the structure of an estimation problem \cite{FGtheory1}. 
Factor graphs are bipartite graphs comprising two kinds of nodes: variable nodes, and factor nodes.
The variables nodes represent the unknown data to be estimated, while factors represent cost functions to be minimized, modeling relationships between the variables.
Each factor node is only connected to the variables that appear in its cost function, giving the FG its characteristic aspect of sparsity. As shown below, by applying this estimation approach to the classical linear bicycle model, good estimation accuracy can be achieved even in the non-linear regions of the tyre behaviour compared to the celebrated Kalman filtering, while preserving the inherent simplicity of the linear model and the use of few parameters (i.e., the front and rear cornering stiffness). 

The paper is organized as follows. In Section \ref{SECT:VehicleModel} the equations relative to the linear bicycle model are recalled and reorganized in order to directly obtain the side slip angle. Section \ref{SECT:Estimation} introduces the estimation problem with the use of FGs, applied to the case at hand with details on its implementation. Results obtained with the proposed method are evaluated by means of real data gathered in the Stanford database \cite{StanfordData} in Section \ref{SECT:Results}. Finally, conclusion are drawn in Section \ref{SECT:Conclusions}.
 
\section{Vehicle dynamic model}
\label{SECT:VehicleModel}

\begin{figure}[h!]
\centering
\includegraphics[width=0.3\textwidth,angle=-10]{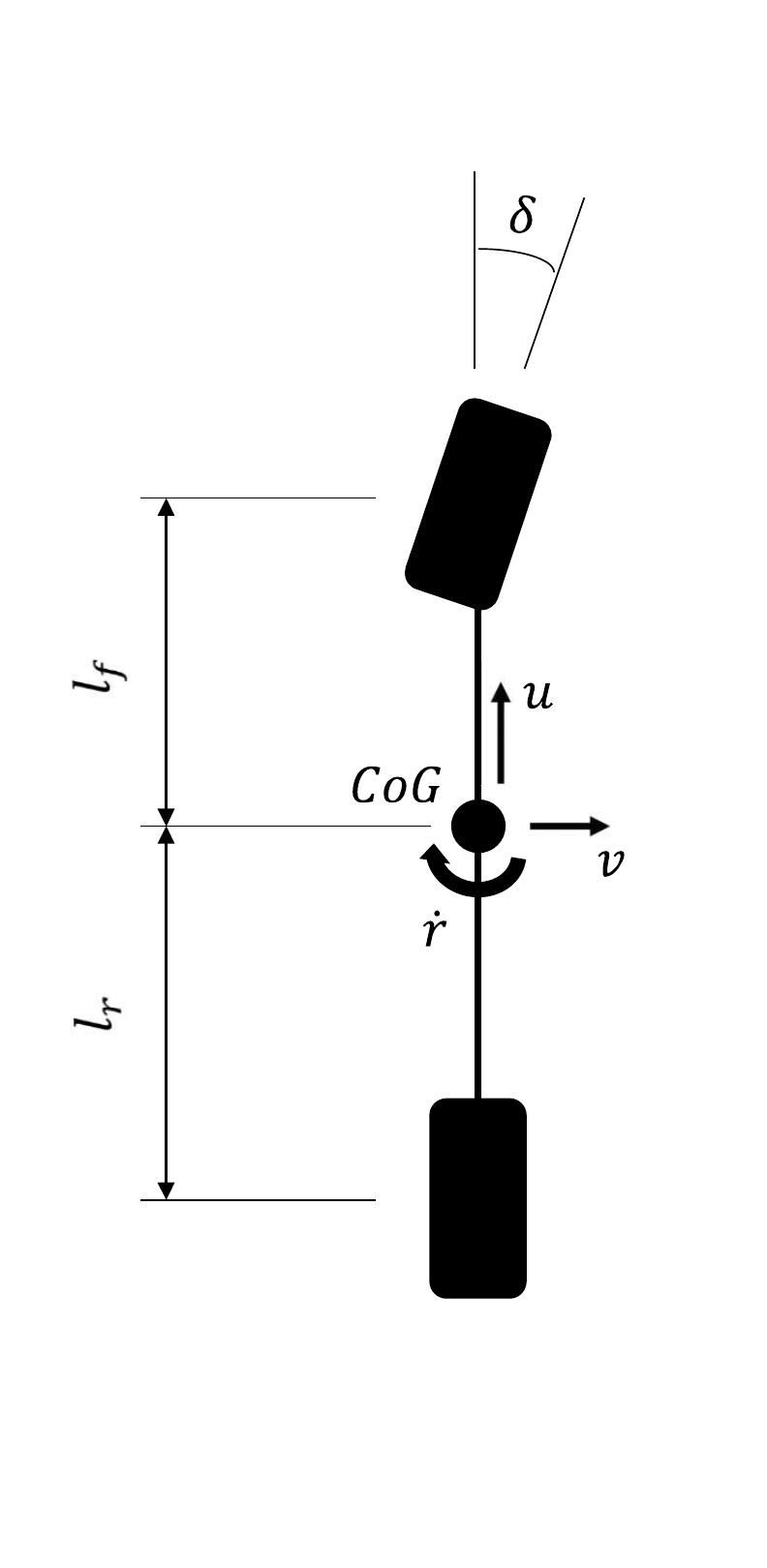}
\caption{single-track model}
\label{FIG:BicycleModel}
\end{figure}

Lateral dynamic represents a basic challenge in vehicle behavior, because lateral velocity is usually not directly measurable for practical and/or economic reasons. Nonetheless, it represents a fundamental information, since it influences the entire vehicle dynamics, especially the direction of the motion. As a matter of fact, the total velocity of the vehicle Center of Gravity (CoG) is the vectorial sum of the longitudinal and lateral velocities $u$ and $v$ respectively. Figure \ref{FIG:BicycleModel} shows the model used in this paper for dealing with the vehicle lateral behavior. It is known as ``Bicycle'' or ``Single-Track'' model and it is based on the following simplifications \cite{LinBicMod}: equal internal and external dynamics so that tires of the same axle can be fused, leading to the model's bicycle-like appearance; linear range of the tires, thus lateral forces are a linear function of slip angles by means of a specific coefficient named Cornering Stiffness; rear-wheel drive; negligible motion resistance; small angle approximation to preserve the linearity of the model; constant longitudinal velocity $u$, that is guaranteed within a small time-step $\Delta t$. This model is characterized by two Degrees of Freedom (DoFs), namely the vehicle lateral velocity $v$ and the yaw rate $r$. Hence, two equations of motion are sufficient to fully describe its behavior over time:
\begin{eqnarray}
&\dot{v} = -\lp\dfrac{C_f+C_r}{mu}\rp v - \lp \dfrac{C_fl_f-C_rl_r}{mu}+u \rp r + \dfrac{C_f\delta}{m} \nonumber \\ \nonumber \\
& \dot{r} = -\lp \dfrac{C_fl_f-C_rl_r}{J_zu} \rp v - \lp \dfrac{C_fl_f^2+C_rl_r^2}{J_zu} \rp + \dfrac{C_fl_f\delta}{J_z}
\label{EQ:DynSys0}
\end{eqnarray}
where $m$ is the vehicle mass, $J_z$ the moment of inertia with respect the yaw axis, $C_f$ and $C_r$ the front and rear cornering stiffness respectively, $l_f$ and $l_r$ the distance of the CoG from the front and rear axle and $\delta$ the front steering angle. 

Usually, the side slip angle $\beta$ is of interest in the study of lateral dynamic, in place of the lateral velocity $v$, therefore one can consider as DoFs $\beta$ and $r$ in place of the couple $v$ and $r$. From Figure \ref{FIG:BicycleModel}, the relation between $v$ and $\beta$ is the following:
\begin{equation}
\beta = \arctan \lp \dfrac{v}{u} \rp \approx \dfrac{v}{u}
\label{EQ:beta}
\end{equation}
where the approximation comes from the assumption of small angles, as stated above. Therefore:
\begin{equation}
v = \beta u \qquad \text{and} \qquad \dot{v} = \dot{\beta}u + \beta\cancelto{0} {\dot{u}} \approx \dot{\beta}u
\end{equation}
considering the constancy of $u$ within each time-step $\Delta t$. The equations (\ref{EQ:DynSys0}) becomes:
\begin{eqnarray}
&\dot{\beta} = -\lp\dfrac{C_f+C_r}{mu}\rp \beta - \lp \dfrac{C_fl_f-C_rl_r}{mu^2}+1 \rp r + \dfrac{C_f\delta}{mu} \nonumber \\ \nonumber \\
& \dot{r} = -\lp \dfrac{C_fl_f-C_rl_r}{J_z} \rp \beta - \lp \dfrac{C_fl_f^2+C_rl_r^2}{J_zu} \rp + \dfrac{C_fl_f\delta}{J_z}
\label{EQ:DynSys1}
\end{eqnarray}

By considering equations (\ref{EQ:DynSys1}), $\beta$ and $r$ are variables of the vehicle lateral dynamics and the other terms are considered known parameters.

It is worth noting that tha yaw rate $r$ can be directly measured by means of a vertical gyroscope, instead the side slip angle $\beta$ usually has to be estimated. Although also $\beta$ can be measured via GPS and equation (\ref{EQ:beta}), often this type of measurement is unstable and unreliable due to the presence of shady areas of the GPS signal, such as tunnels or mountains. Hence, the need to estimate indirectly the side slip angle. To achieve this estimation the lateral acceleration $a_y$ is considered as a further measurement. Both $a_y$ and $r$ are generally already available onboard common vehicles via the ESP system. Therefore, the following set of measurements is here considered:
\begin{eqnarray}
&\dot{\varphi} = r \nonumber \\
&a_y = \dot{v} + ur = u(\dot{\beta}+r)
\label{EQ:Meas1}
\end{eqnarray}

As a matter of practicality, the equations (\ref{EQ:DynSys1}) and (\ref{EQ:Meas1}) in their continuous-time form need to be converted in the following discrete-time representation:
\begin{subequations}
\begin{equation}
\beta_k = \beta_{k-1} + \Delta t \left[ -\dfrac{C_f+C_r}{mu_{k-1}}\beta_{k-1} - \lp \dfrac{C_fl_f-C_rl_r}{mu_{k-1}^2}+1 \rp r_{k-1} + \dfrac{C_f\delta_{k-1}}{mu_{k-1}} \right]
\end{equation}
\\
\begin{equation}
r_k = r_{k-1} + \Delta t \lp -\dfrac{C_fl_f-C_rl_r}{Jz}\beta_{k-1} - \dfrac{C_fl_f^2+C_rl_r^2}{Jzu_{k-1}}r_{k-1} + \dfrac{C_fl_f\delta_{k-1}}{Jz} \rp
\end{equation}
\\
\begin{equation}
\dot{\varphi}_k = r_k
\end{equation}
\\
\begin{equation}
a_{yk} = -\dfrac{C_f+C_r}{m}\beta_k - \dfrac{C_fl_f-C_rl_r}{mu_k}r_k + \dfrac{C_k\delta_k}{m}
\end{equation}
\label{EQ:DiscrEqs}
\end{subequations}
The above equations are obtained from the forward Euler integration in the time step $\Delta t = t_k - t_{k-1}$.

\section{Factor Graph for vehicle lateral dynamics}
\label{SECT:Estimation}

\subsection{The estimation problem}

Discrete-time equations (\ref{EQ:DiscrEqs}) refer to the vehicle system and to the measurements gathered for a correct estimate of the DoFs, in particular for the estimation of $\beta$, since $r$ is directly measured. The most common way followed throughout the literature relies on a state-space representation of the above equations, writing two compacted matrix equations: one for the propagation of the model over time and one relating to the measures, both as a function of suitable state variables. Therefore, a recursive routine, often based on Kalman filtering, is performed to achieve the correct estimate of the unknown state variables \cite{LeanzaReinaMantriota}.

\begin{figure}[h!]
\centering
\includegraphics[width=0.7\textwidth]{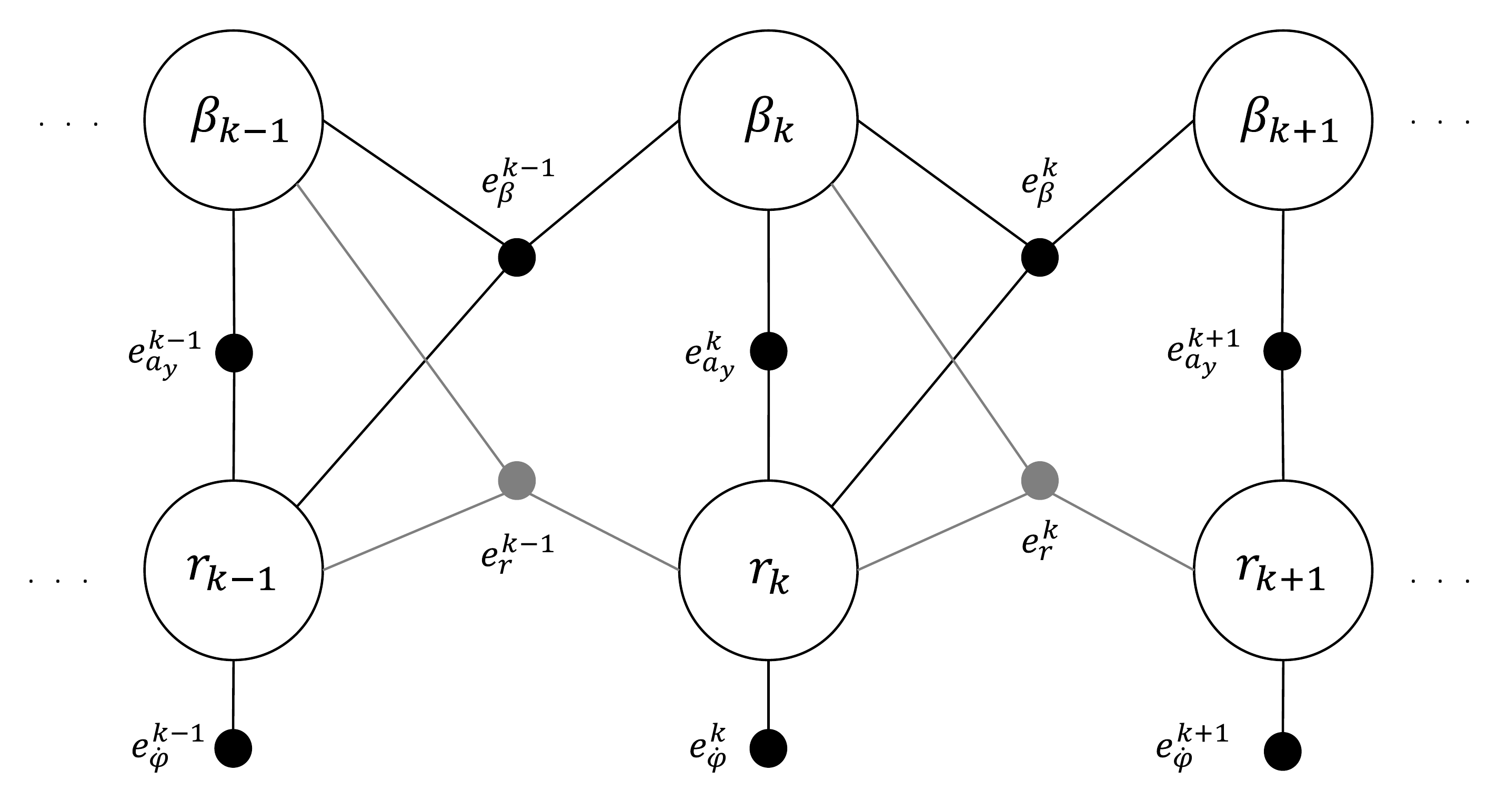}
\caption{FG relative to the linear bicycle model}
\label{FIG:FGlbm}
\end{figure}

In this paper, an alternative approach is suggested for the estimation of the unknown variables, without passing through the state-space form, namely the Factor Graphs (FGs), a method belonging to the family of probabilistic graphical models. The objective is to minimize given cost functions in each time step. The FG relative to the problem at hand is displayed in Figure \ref{FIG:FGlbm} for a generic time-step $k$. Each factor (filled circle in the figure) is connected with the unknown variables involved in the cost function that the factor minimizes. Cost functions come directly from equations (\ref{EQ:DiscrEqs}):
\begin{subequations}
\begin{equation}
e_\beta^{k-1} = \beta_k - \beta_{k-1} - \Delta t \left[ -\dfrac{C_f+C_r}{mu_{k-1}}\beta_{k-1} - \lp \dfrac{C_fl_f-C_rl_r}{mu_{k-1}^2}+1 \rp r_{k-1} + \dfrac{C_f\delta_{k-1}}{mu_{k-1}} \right]
\label{EQ:e_beta}
\end{equation}
\\ 
\begin{equation}
e_r^{k-1} = r_k - r_{k-1} - \Delta t \lp -\dfrac{C_fl_f-C_rl_r}{Jz}\beta_{k-1} - \dfrac{C_fl_f^2+C_rl_r^2}{Jzu_{k-1}}r_{k-1} + \dfrac{C_fl_f\delta_{k-1}}{Jz} \rp
\label{EQ:e_r}
\end{equation}
\\
\begin{equation}
e_{\dot{\varphi}}^k = \dot{\varphi} - r_k
\label{EQ:e_dphi}
\end{equation}
\\
\begin{equation}
e_{a_y}^k = a_{yk} + \dfrac{C_f+C_r}{m}\beta_k + \dfrac{C_fl_f-C_rl_r}{mu_k}r_k - \dfrac{C_f\delta_k}{m}
\label{EQ:e_ay}
\end{equation}
\label{EQ:errors}
\end{subequations}

The cost functions $e_\beta^{k-1}$ and $e_r^{k-1}$ are handled, respectively, by the black and gray ternary factors (Figure \ref{FIG:FGlbm}). On the other hand, the unary and binary factors are referred to the measurements: the unary factor minimizes the difference between the actual yaw rate $\dot{\varphi}$ and the associated unknown $r$, whilst the binary factor minimizes the difference between the actual lateral acceleration $a_y$ and the estimated one.

In principle, the value of the error functions (\ref{EQ:errors}) has to be equal to zero, but actually an uncertainty is associated to each factor due to the stochastic nature of the estimation problems. First, measurements are collected through the use of sensors with inherent precision and accuracy, 
on the other hand, the uncertainties associated to the model include mathematical approximations and uncertainties of the inputs $u$ and $\delta$. 

Without loss of generality, with the assumption of zero-mean Gaussian probability density function (pdf) associated to the uncertainty in each factor, the minimization problem can be reduced to a linear least squares problem. Since the unknown variables $\mathbf{X}_k$ to minimize at the $k-th$ instant are $\left[\beta_k \quad r_k\right]^T$, equations (\ref{EQ:errors}) can be written in compact matrix form as:
\begin{equation}
\begin{bmatrix}
e_\beta \\ e_r \\ e_{\dot{\varphi}} \\ e_{a_y} 
\end{bmatrix}_k =
\mathbf{H}_k \mathbf{X}_k + \mathbf{C}_k \mathbf{h}_k
\end{equation}
where 
\begin{eqnarray*}
&\mathbf{H}_k = 
\begin{bmatrix}
-\lp 1-\Delta t\dfrac{C_f+C_r}{mu_{k-1}}\rp && \Delta t\lp \dfrac{C_fl_f-C_rl_r}{mu_{k-1}^2}+1\rp && 1 && 0 \\ \\
\Delta t\dfrac{C_fl_f-C_rl_r}{J_z} && -\lp1-\Delta t\dfrac{C_fl_f^2+C_rl_r^2}{Jzu_{k-1}}\rp && 0 && 1 \\ \\
0 && 0 && 0 && -1 \\ \\
0 && 0 && \dfrac{C_f+C_r}{m} && \dfrac{C_fl_f-C_rl_r}{mu_k}
\end{bmatrix} \\ \\ \\
&\mathbf{C}_k = \text{diag} \lp \dfrac{C_f}{mu_{k-1}} \quad \dfrac{C_fl_f}{J_z} \quad 0 \quad -\dfrac{C_f}{m} \rp \\ \\ \\
&\mathbf{h}_k = \left[ \delta_{k-1} \quad \delta_{k-1} \quad  0 \quad  \delta_k \right]^T
\end{eqnarray*}
By defining the \textit{state update vector} as $\bm{\Delta}_k := \mathbf{X}_k - \mathbf{X}_{k-1}$, the next equation holds:
\begin{equation}
\mathbf{H}_k\mathbf{X}_k = \mathbf{H}_k\bm{\Delta}_k + \mathbf{H}_k\mathbf{X}_{k-1}
\end{equation}
The target is the estimation of the state update vector $\bm{\Delta}_k$ via least squares. It is important to remark that the problem at hand is linear, therefore the following Weighted Least Squares (WLS) problem is set:
\begin{equation}
\hat{\bm{\Delta}} = 
\operatorname*{argmin}_{\bm{\Delta}} \operatorname*{\sum}_k 
\big{\lVert} 
\mathbf{H}_k\bm{\Delta}_k - \lp \z_k - \mathbf{H}_k\mathbf{X}_{k-1} - \mathbf{C}_k\mathbf{h}_k \rp
\big{\lVert}_{\Q_k}^2
\end{equation}
with $\z_k$ the $k-th$ vector of measures, $\Q_k$ the error covariance matrix and the parenthesis $\lp \z_k - \mathbf{H}_k\mathbf{X}_{k-1} - \mathbf{C}_k\mathbf{h}_k \rp$ is the \textit{prediction error}. The argument of the sum is the Mahalanobis norm:
\begin{equation}
\left[ \Q_k^{-\frac{1}{2}} \left[ \mathbf{H}_k\bm{\Delta}_k - \lp \z_k - \mathbf{H}_k\mathbf{X}_{k-1} - \mathbf{C}_k\mathbf{h}_k \rp \right] \right]^T
\left[ \Q_k^{-\frac{1}{2}} \left[ \mathbf{H}_k\bm{\Delta}_k - \lp \z_k - \mathbf{H}_k\mathbf{X}_{k-1} - \mathbf{C}_k\mathbf{h}_k \rp \right] \right]
\end{equation}
and from the next replacements:
\begin{subequations}
 \begin{equation}
 \Q_k^{-\frac{1}{2}}\mathbf{H}_k = \A_k
 \end{equation}
 \begin{equation}
 \Q_k^{-\frac{1}{2}}\lp \z_k - \mathbf{H}_k\mathbf{X}_{k-1} - \mathbf{C}_k\mathbf{h}_k \rp = \bk_k
 \label{EQ:b_k}
 \end{equation}
\end{subequations} 
one obtains the following simple least squares problem:
\begin{equation}
\hat{\bm{\Delta}} = 
\operatorname*{argmin}_{\bm{\Delta}}
\big{\lVert} 
\A\bm{\Delta} - \bk
\big{\lVert}_2^2 = \lp\ \A^T\A \rp^{-1}\A^T\bk
\label{EQ:lsq}
\end{equation}
where $\A$ is a large matrix collecting all matrices $\A_k$. From the calculation of the state update vector: $\hat{\mathbf{X}}_k = \hat{\mathbf{X}}_{k-1} + \hat{\bm{\Delta}}_k$, where $\A$, $\bk$ and $\bm{\Delta}$ grow over time.

\subsection{Implementation}

\begin{figure}[h!]
\centering
\includegraphics[width=0.7\textwidth]{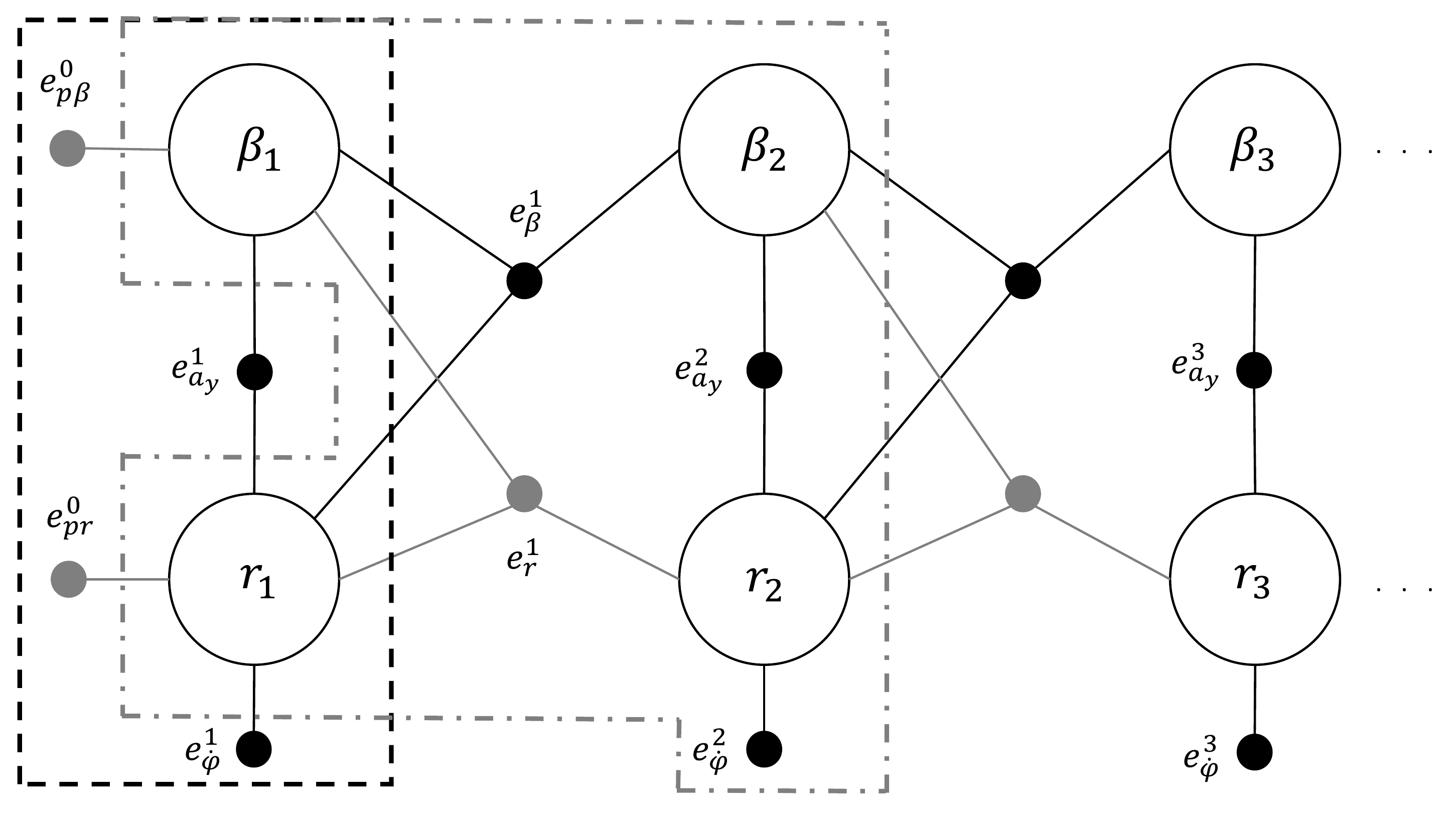}
\caption{FG relative to the first three steps of the estimation problem: into the black dashed window the first step with priors and into the gray dash-dot one the generic step, with involved factors}
\label{FIG:FGwP}
\end{figure}

Figure \ref{FIG:FGwP} displays the first three steps for the estimation problem at hand. The first step, surrounded by the closed dashed black line, includes two priors factors (the unitary gray ones) in place of the dynamic factors. The priors factors minimize the difference between the first estimates and the initial guess of the unknown variables, enforcing known initial conditions; therefore reliable initial guess is assumed, with values very close to the correct ones.  Their error functions are:
\begin{subequations}
\begin{equation}
e_{p\beta}^0 = \beta_1 - \beta_0
\end{equation}
\begin{equation}
e_{pr}^0 = r_1 - r_0
\end{equation}
\end{subequations}

The other steps follow the logic of the second one, contained in the closed dash-dot gray line. This closed line encloses four unknown variables and four factors: two belonging to the dynamic model and two for the measures. By looking at the first dynamic step enclosed in the gray line, the dynamic model is managed by the ternary factors $e_\beta^1$ and $e_r^1$ for the estimation of the side slip angle and the yaw rate respectively, while factors $e_{a_y}^2$ and $e_{\dot{\varphi}}^2$ introduce new measurements and minimize the difference between what is effectively measured and the what is expected from the model.

In order to understand how to move from the graphical representation of factors to the minimization problem, the $\A$ matrix and vectors $\bk$ and $\bm{\Delta}$ are shown for the first three steps of Figure \ref{FIG:FGwP}. 

\begin{table}[h!]
\centering
\caption{$\A$ matrix and $\bk$ vector for the first three steps of estimation}
\vspace*{0.3cm}
\begin{tabular}{|c|cccccc|c|c|}
\cline{1-7} \cline{9-9}
\textbf{Factors} & $\Delta_1$ & $\Delta_2$                     & $\Delta_3$ & $\Delta_4$                     & $\Delta_5$ & $\Delta_6$ &  & $\bk$    \\ \cline{1-7} \cline{9-9} 
p1      & $A_{1,1}$  & \multicolumn{1}{c|}{}          &            &                                &            &            &  & $b_1$    \\
p2      &            & \multicolumn{1}{c|}{$A_{2,2}$} &            &                                &            &            &  & $b_2$    \\
m1      &            & \multicolumn{1}{c|}{$A_{3,2}$} &            &                                &            &            &  & $b_3$    \\
m2      & $A_{4,1}$  & \multicolumn{1}{c|}{$A_{4,2}$} &            &                                &            &            &  & $b_4$    \\ \cline{2-5}
d1      & $A_{5,1}$  & $A_{5,2}$                      & $A_{5,3}$  & \multicolumn{1}{c|}{}          &            &            &  & $b_5$    \\
d2      & $A_{6,1}$  & $A_{6,2}$                      &            & \multicolumn{1}{c|}{$A_{6,4}$} &            &            &  & $b_6$    \\
m3      &            &                                &            & \multicolumn{1}{c|}{$A_{7,4}$} &            &            &  & $b_7$    \\
m4      &            &                                & $A_{8,3}$  & \multicolumn{1}{c|}{$A_{8,4}$} &            &            &  & $b_8$    \\ \cline{2-7}
d3      &            & \multicolumn{1}{c|}{}          & $A_{9,3}$  & $A_{9,4}$                      & $A_{9,5}$  &            &  & $b_9$    \\
d4      &            & \multicolumn{1}{c|}{}          & $A_{10,3}$ & $A_{10,4}$                     &            & $A_{10,6}$ &  & $b_{10}$ \\
m5      &            & \multicolumn{1}{c|}{}          &            &                                &            & $A_{11,6}$ &  & $b_{11}$ \\
m6      &            & \multicolumn{1}{c|}{}          &            &                                & $A_{12,5}$ & $A_{12,6}$ &  & $b_{12}$ \\ \cline{1-7} \cline{9-9} 
\end{tabular}
\label{TAB:A&b}
\end{table}

In Table \ref{TAB:A&b} $\A$ matrix and $\bk$ vector are reported, instead  the $\bm{\Delta}$ vector for the first three steps is that in equation (\ref{EQ:Delta}).

\begin{equation}
\begin{bmatrix}
\Delta_1 \\ \Delta_2 \\ \Delta_3 \\ \Delta_4 \\ \Delta_5 \\ \Delta_6
\end{bmatrix} = 
\begin{bmatrix}
\beta_1-\beta_0 \\ r_1-r_0 \\ \beta_2-\beta_1 \\ r_2-r_1 \\ \beta_3-\beta_2 \\ r_3-r_2
\end{bmatrix}
\label{EQ:Delta}
\end{equation}

By looking at the matrix $\A$ in Table \ref{TAB:A&b}, it is composed of blocks each one corresponding to a single time-step of the FG, as shown in Figure \ref{FIG:FGwP}. In fact, the first block corresponds to the black dashed window containing the two priors, two measures and the first two unknown variables $\beta_1$ and $r_1$. Hence, the whitened matrix $\A_1$ is a $4 \times 2$ matrix, where each row is associated to a specific factor of Figure \ref{FIG:FGwP}. The whitened matrix $\A_2$ is composed of a square matrix containing the first and second block together and so forth. Vectors $\bk$ and $\bm{\Delta}$ grow accordingly. It is worth noting that the resulting matrix is a sparse matrix, with elements concentrated near the diagonal. This characteristic can be exploited in the minimization problem of equation (\ref{EQ:lsq}). For further details on the elements of matrix $\A$ and vector $\bk$ in Table \ref{TAB:A&b}, the interested reader can refer to Appendix.

Matrix $\Q_k$ collects the uncertainty associated to each factor. Therefore, $\Q_k^{-\frac{1}{2}}$ contains the weights of the factors, representing the tuning parameters of this estimation problem. As a matter of fact, the greater the confidence given to a factor, the higher the values associated with its weight and then the lower the corresponding standard deviation. In this case, by looking at equations (\ref{EQ:errors}) it is clear that the factor associated to the yaw rate measurement has a large weight, being precisely measured, therefore, the standard deviation associated to the error of equation (\ref{EQ:e_dphi}) is very small. In this work a value of $\sigma_{\dot{\varphi}}=10^{-8}$ [rad/s] has been assumed. Instead, the error associated to the equation (\ref{EQ:e_ay}) relative to the lateral acceleration measurement is taken larger because a linear model is compared with actual lateral acceleration, therefore a smaller weight is associated to that factor, by assuming a $\sigma_{a_y}=10^{-2}$ $[\text{m}/\text{s}^2]$. Regarding the error relative to the model equations, a high weight is associated to the corresponding factors for the well-known reliability of the model, with standard deviations equal to $\sigma_\beta = 10^{-5}$ [rad] and $\sigma_r = 10^{-4}$ [rad/s] for the side slip angle model and yaw rate model respectively. 

During vehicle motion, a huge amount of data can be collected, and performing a batch estimation on a too long time window would be impractical. Therefore, a fixed-lag-smoother is applied by considering a sliding window, which contains a fixed number $M$ of samples. Hence, the minimization is achieved on these samples and the most reliable estimate is retained as a set of priors for the next estimation. The length of the sliding window is another tuning parameter of this estimation approach. In this work, a window encasing $M=5$ samples is heuristically found to be a a good trade-off between computational speed and goodness of estimation. By looking at Figure \ref{FIG:FGwP}, this window will be composed by one closed dashed black line with the priors consisting of the previous estimate (or of the initial conditions for the first window) and by five closed dash-dot gray line, where the indices of every factors change accordingly with the time step. Of course, the $\A$ matrix will be composed by six blocks following the rationale of Table \ref{TAB:A&b} and vectors $\bk$ and $\bm{\Delta}$ grow accordingly.

\section{Results}
\label{SECT:Results}

This section collects results obtained with the proposed approach and evaluated by means of real data acquired by an instrumented Ferrari 250 LM Berlinetta GT and made publicly available by Center for Automotive Research at Stanford \cite{StanfordData}. A Global Navigation
Satellite System (GNSS)-aided Inertial Navigation System provides overall vehicle body motion resulting in centimeter-level position accuracy and slip-angle direct measurement. The 2014 Targa Sixty-Six event served as the data collection venue that took place at the Palm Beach International Raceway, a 3.3 km-long track featuring of 10 turns and a 1 km straight.

\begin{figure}[hbtp]
\centering
\includegraphics[width=0.65\textwidth]{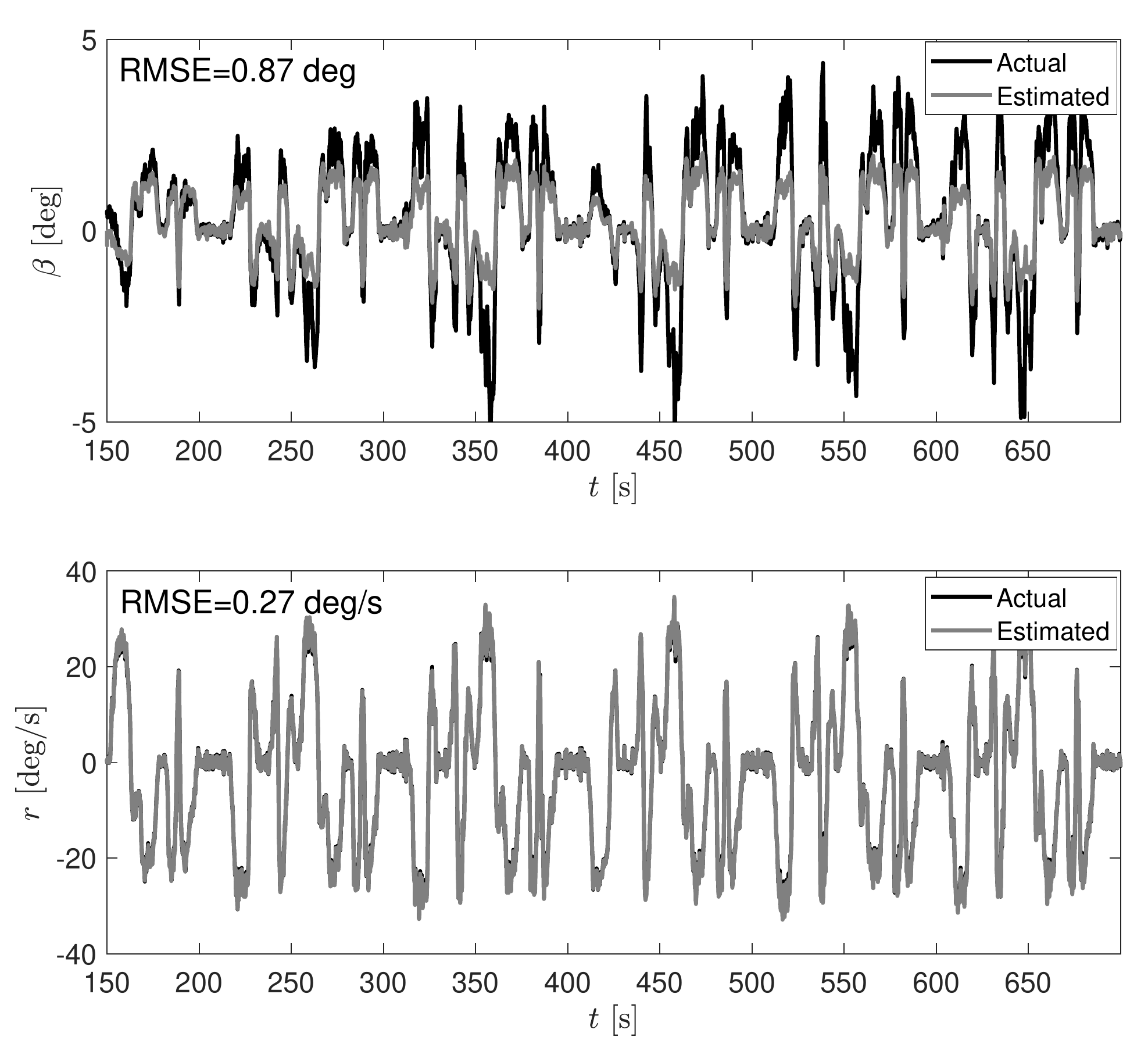}
\caption{Vehicle side slip angle $\beta$ estimate at the top and yaw rate $r$ at the bottom by using a KF-based observer applied to the linear bicycle model}
\label{FIG:LinearKF1}
\end{figure}

For a thoroughly understanding of the advantages of the proposed method, the results obtained with the celebrated linear KF are first shown, which are well known in the automotive field. 
For the sake of brevity, the KF equations and the state-space form of the problem at hand is here omitted, since it is widespread in the literature, e.g., \cite{ReinaMessina}. 
Figure \ref{FIG:LinearKF1} shows the estimation of the vehicle side slip angle $\beta$ and yaw rate $r$ obtained via linear Kalman filtering during a full lap of the Palm Beach International circuit. 
As expected, the unknown state associated to the yaw rate is estimated very well, since it is directly measured. In fact only the sensor noise is filtered out leading to an RMSE value of 0.27 deg/s.
Note as well the relatively good estimation of $\beta$ (RMSE of 0.87 deg) despite the approximation of assuming a linearized model.
This is explained by the relatively small range of $\beta$ values observed in practice and, in particular, for this race dataset.


\begin{figure}
\centering
\includegraphics[width=0.65\textwidth]{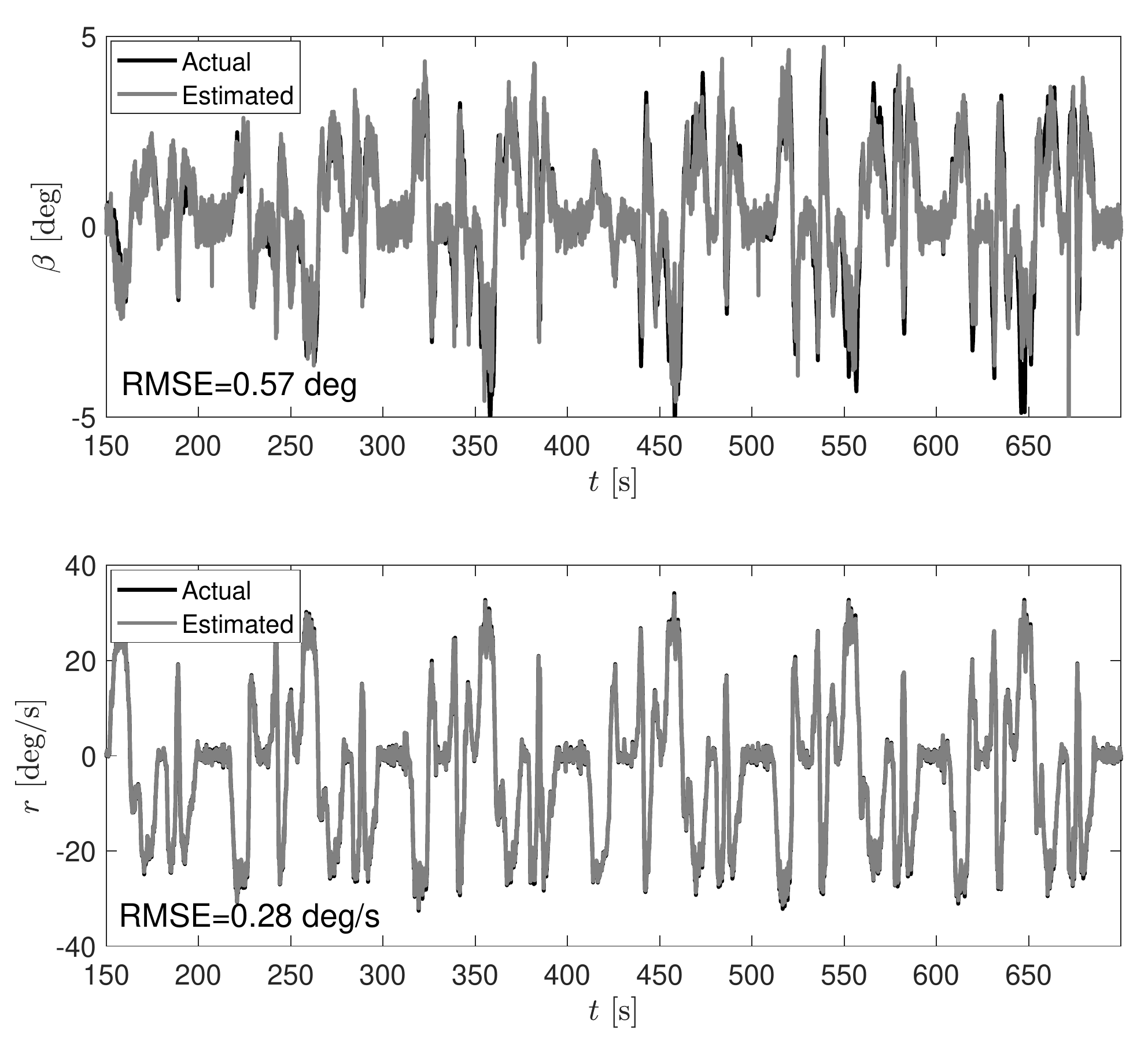}
\caption{Vehicle side slip angle $\beta$ estimate at the top and yaw rate $r$ at the bottom by using FG applied to the linear bicycle model}
\label{FIG:FG1}
\end{figure}

In Figure \ref{FIG:FG1} the estimation of the side slip angle $\beta$ and yaw rate $r$ is shown, but this time obtained from the FG-based observer. As seen from the figure, the estimator can track $\beta$ more closely and even for higher values than KF notwithstanding it is based on a linear model. A corresponding RMSE of 0.57 deg is achieved that is a 34\% improvement over the KF implementation.
Note that numerical estimators used for the FG are capable of optimally handle strongly non-linear model due to their iterative nature, but in this particular case this advantage is not exploited due to the linearity of the model. In fact, the adopted nonlinear solver (Gauss-Newton) only runs a single iteration before detecting it reached the optimum. 

\begin{figure}[hbtp]
\centering
\includegraphics[width=0.6\textwidth]{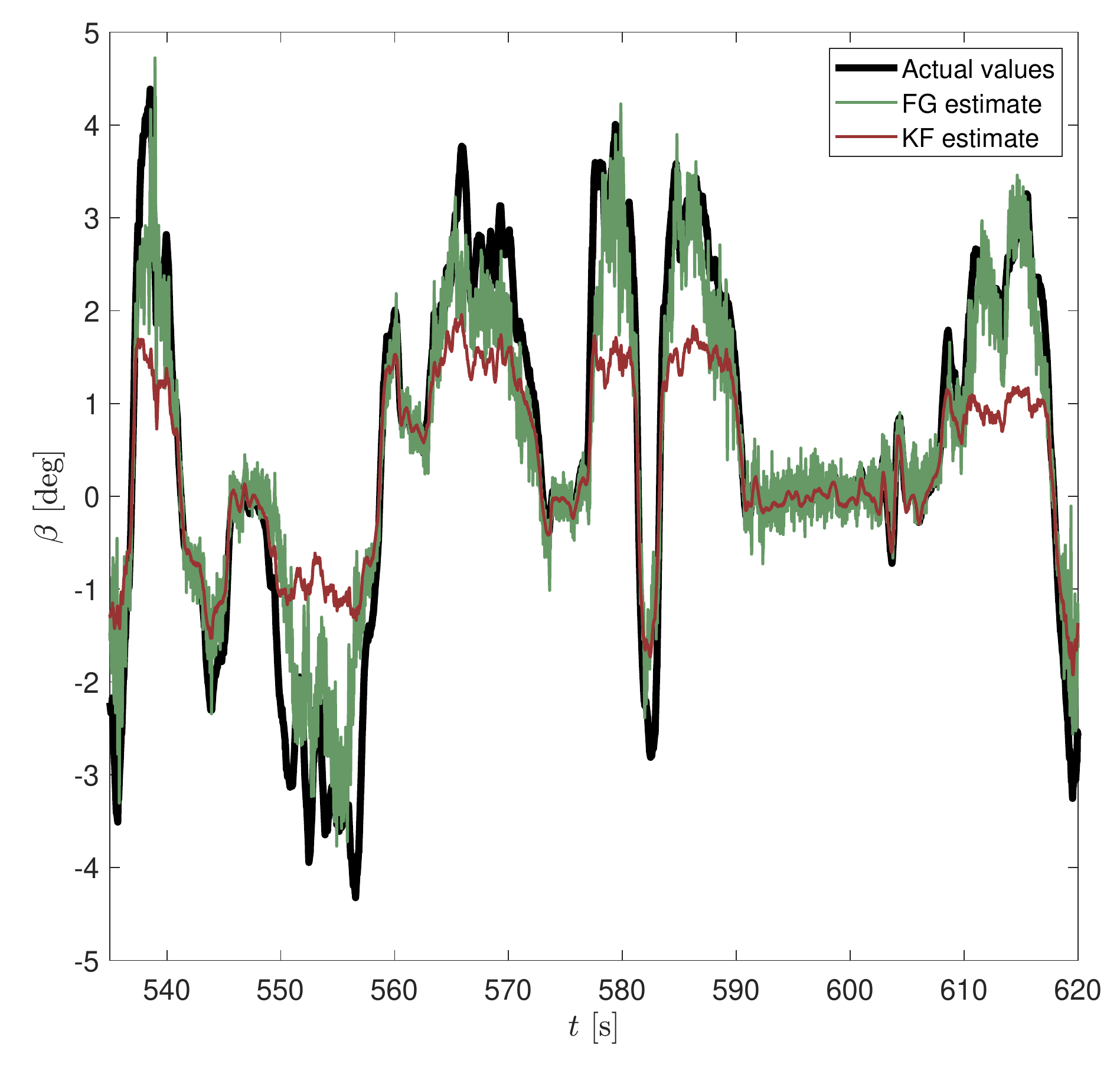}
\caption{Vehicle side slip angle $\beta$ estimation by considering both KF and FG estimators for small values of the side slip angle}
\label{FIG:zoom1}
\end{figure}

On the other hand, one fundamental difference between the FG solver and the KF is the use of a sliding-window estimator for the former, which is then capable of smoothing out sensor noise 
much more effectively than KF is able to by sequentially processing time steps one by one. 
However, for very small $\beta$ angles (see detailed view in Figure \ref{FIG:zoom1}), 
KF seems to provide a less noisy output. This effect may be explained by the use of different
tuning parameters in both approaches. 


\begin{figure}[hbtp]
\centering
\includegraphics[width=0.6\textwidth]{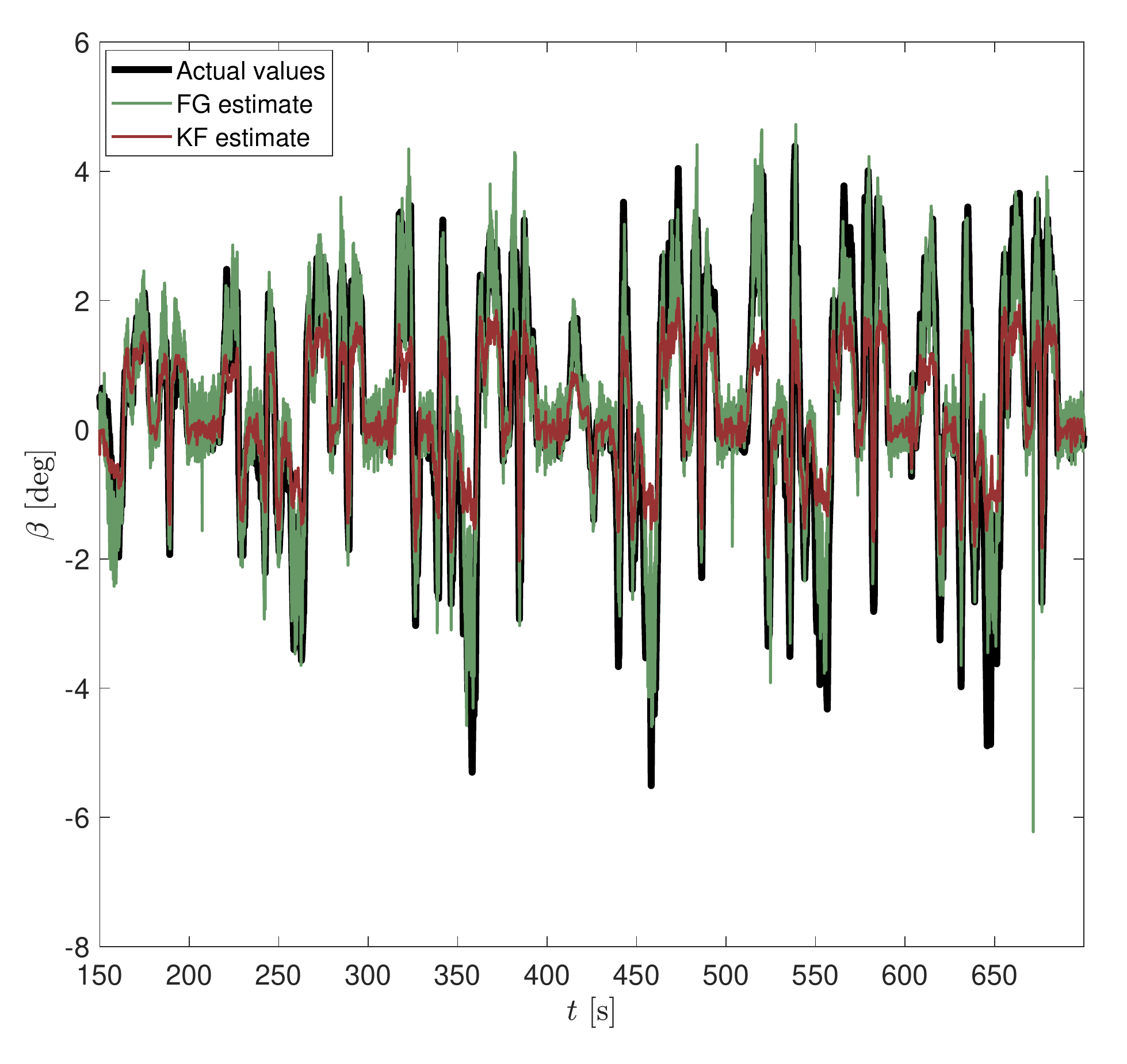}
\caption{Vehicle side slip angle $\beta$ estimation by considering both KF and FG estimators with a window of 5 samples}
\label{FIG:comparison1}
\end{figure}

\begin{figure}[hbtp]
\centering
\includegraphics[width=0.6\textwidth]{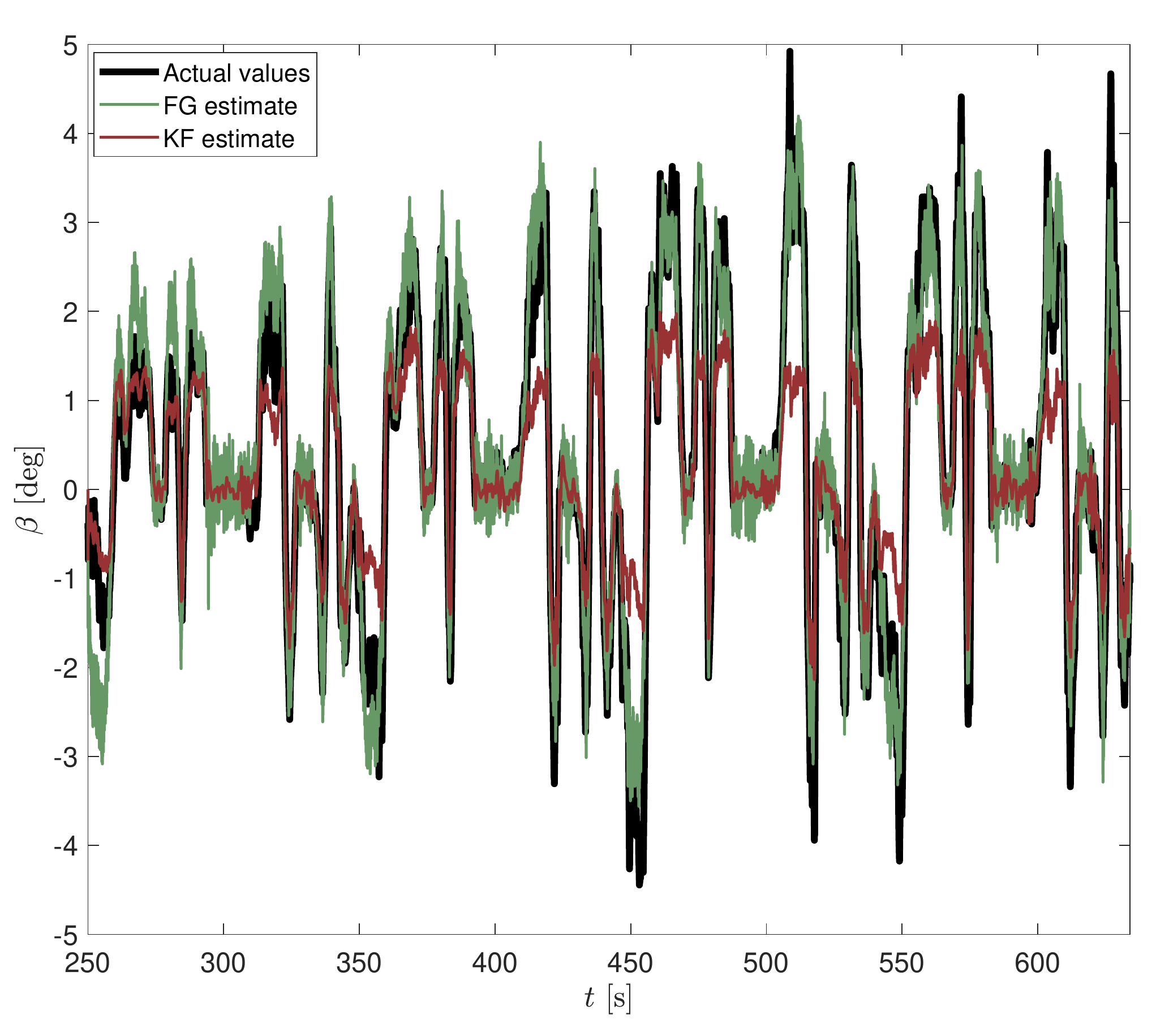}
\caption{Vehicle side slip angle $\beta$ estimation by considering both KF and FG estimators with a window of 5 samples, for another set of real data}
\label{FIG:comparison2}
\end{figure}

For completeness, another experiment performed on a different set of real data gathered in a different race session is displayed in Figure \ref{FIG:comparison2}. Again, the KF shows the best performance in the linear region characterized by small $\beta$ values, but fails for higher values of side slip angles. Vice versa, the FG is noisier for small $\beta$ values but well estimates the actual side slip angle for higher values, that are the most interesting one for the vehicle handling and passengers safety.

\begin{figure}[hbtp]
\centering
\includegraphics[width=0.6\textwidth]{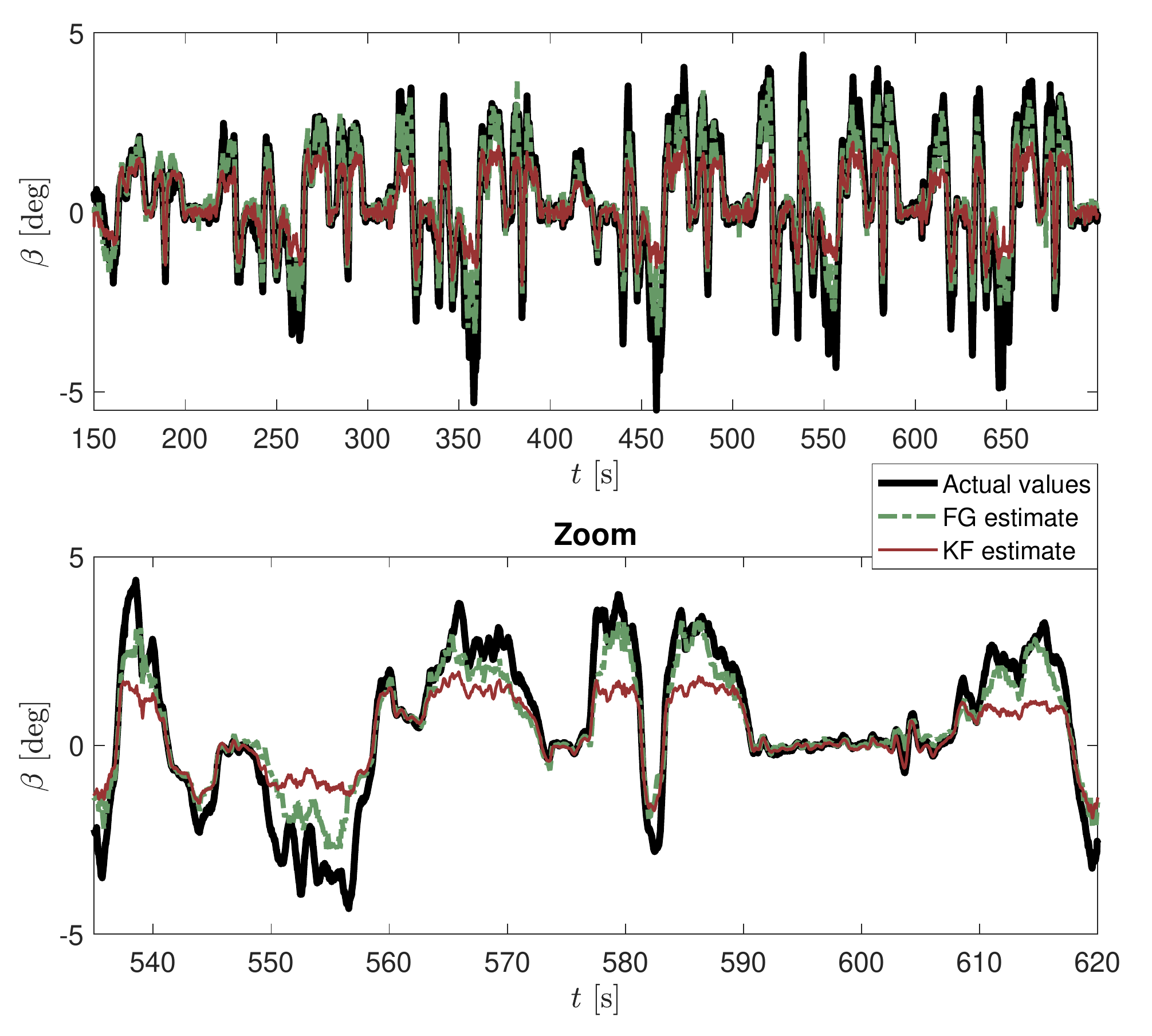}
\caption{Vehicle side slip angle $\beta$ estimation by considering both KF and FG batch estimators}
\label{FIG:batch}
\end{figure}

In order to show the potential of the proposed method, the batch estimation is reported in Figure \ref{FIG:batch} on the total number of samples acquired. As one can see, both the advantages of KF and FG with sliding window method are obtained, consisting of a good estimation for high values of $\beta$ beyond the linear region and a very smooth estimate especially in the linear one, similarly to the KF, where a better visualization has been achieved at the bottom of the figure in a shorter time length.

%

\section{Conclusions}
\label{SECT:Conclusions}

In this paper an estimator grounded on the Factor Graph theory is applied for the first time for estimating the side slip angle of a vehicle during motion. A linear single-track model is considered in this research as the vehicle model the estimator. The performance of the proposed estimation approach are evaluated by using real data and contrasted with those obtained from the standard Kalman Filter. It was demonstrated that the proposed method estimates accurately the side slip angle even when its value exceeds the linearity range, which represents the more critical situation in terms of vehicle handling and passenger safety, and thus the knowledge of the correct $\beta$ value becomes more important. On the other hand, for small values of the side slip angle, the Kalman filtering preserves its superiority, as expected. Nonetheless, a batch estimation leads to a good estimate of the side slip angle with a level of noise similar to that achieved by using the KF. However, it is worth pointing out that higher values of the side slip angle can be correctly estimated by keeping a linear bicycle model with very few parameters with respect to other more complex vehicle models, representing an important aspect in terms of practical implementation on real commercial vehicles.

Next steps will take into account FGs applied on more complex (non-linear) vehicle models, in order to investigate the effectiveness and the ability to achieve even more accurate estimates and also other aspects related to vehicle dynamics that cannot be represented with the bicycle model.

\vspace{6pt} 






\section*{Funding}
The financial support of the projects: Agricultural inTeroperabiLity and Analysis System (ATLAS), H2020
(Grant No. 857125), and multimodal sensing for individual plANT phenOtypiNg in agrIculture rObotics (ANTONIO), ICT-AGRI-FOODCOFUND(GrantNo. 41946) is gratefully acknowledged.



%


\appendix
\section*{Appendix}

In this Appendix, details about $\A$ matrix and $\bk$ vector of Table \ref{TAB:A&b} are provided, for the convenience of the interested reader. By starting with the $\A$ matrix, two different kinds of block matrix can be found: one associated to the first step with priors and one associated to the others steps with the dynamic model. By recalling that:
\begin{equation}
\Q^{-\frac{1}{2}} =\text{diag}\lp \dfrac{1}{\sigma_\beta}, \quad \dfrac{1}{\sigma_r}, \quad \dfrac{1}{\sigma_{\dot{\varphi}}}, \quad \dfrac{1}{\sigma_{a_y}} \rp 
\end{equation}
the former kind of block matrix is a $4 \times 2$ matrix where:
\begin{equation}
\begin{bmatrix}
A_{1,1} && 0 \\
0 && A_{2,2} \\
0 && A_{3,2} \\
A_{4,1} && A_{4,2} \\
\end{bmatrix} = 
\Q^{-\frac{1}{2}}
\begin{bmatrix}
1 && 0 \\
0 && 1 \\
0 && -1 \\
\dfrac{C_f+C_r}{m} && \dfrac{C_fl_f-C_rl_r}{mu_k} \\
\end{bmatrix}
\label{APDXEQ:PriorA}
\end{equation} 
The longitudinal velocity $u_k$ is that measured at the specific time step on wich the priors are applied. Foe instance, for the first time step with priors associated to initial guess, $u_k=u(k=1)$. The other terms are constant in this work. The latter kind of block matrix is a $4 \times 4$ matrix obtained as in the following:
\begin{equation}
\begin{bmatrix}
A_{1,1} && A_{1,2} && A_{1,3} && 0 \\
A_{2,1} && A_{2,2} && 0 && A_{2,4} \\
0 && 0 && 0 && A_{3,4} \\
0 && 0 && A_{4,3} && A_{4,4}
\end{bmatrix} = 
\Q^{-\frac{1}{2}} ~ \mathbf{H}_k
\label{APDXEQ:DynamicA}
\end{equation}
with $\mathbf{H}_k$ depending of the $k-th$ time step. A smoother with a window comprising of 5 samples has an $\A$ matrix composed by one matrix (\ref{APDXEQ:PriorA}) and five matrices (\ref{APDXEQ:DynamicA}) arranged on the diagonal.

Vector $\bk$ grows accordingly with $\A$ matrix. By looking at equation (\ref{EQ:b_k}), the parentheses is the prediction error between actual value and estimated one. Therefore: 
\begin{itemize}
\item for priors the actual value is the guess and the function is the difference between these two values, hence:
\begin{equation}
b_1 = \lp \beta_0 -\cancel{\beta_0} + \cancel{\beta_0} \quad \rp \big{/}\sigma_\beta
\end{equation}
and $b_2$ follows the same rationale;
\item for dynamic factors the actual value is zero, since it is the difference between the forward value and the same obtained by integrating the differential equation, for instance:
\begin{equation}
b_5 = \left[ 0 - \beta_1 + \lp 1-dt\dfrac{C_f+C_r}{mu_1}\rp \beta_0 - dt\lp \dfrac{C_fl_f-C_rl_r}{mu_1^2}+1\rp r_0 + dt\dfrac{C_f\delta_1}{mu_1} \right]\bigg{/}\sigma_r
\end{equation} 
and $b_6$ follows the same rationale;
\item finally measures follow this rationale (only yaw rate is taken for the sake of brevity):
\begin{equation}
b_3 = \lp \dot{\varphi}_1 + r_0 \rp\big{/}\sigma_{\dot{\varphi}}
\end{equation}
\end{itemize}
Of course, all indices change according to the time step.


\bibliographystyle{plain}
\bibliography{FGautomotiveBiblio.bib}  

\end{document}